\documentclass{article}

\usepackage[cmex10]{amsmath}
\usepackage{amsthm}
\usepackage{spconf}
\usepackage{amssymb}
\usepackage{mathrsfs}
\usepackage{graphicx}
\usepackage{float}
\usepackage{array}
\usepackage{epstopdf}
\usepackage{multirow}

\newcommand{\argmin}{\arg\!\min}
\newcommand{\argmax}{\arg\!\max}
\allowdisplaybreaks[1]	

\begin{document}

\title{ Highly Accurate Palmprint Recognition Using Statistical and Wavelet Features }

\name{Shervin Minaee and AmirAli Abdolrashidi}
\address{ECE Department, NYU Polytechnic School of Engineering, NY, USA}

\maketitle

\begin{abstract}

Palmprint is one of the most useful physiological biometrics that can be used as a powerful means in personal recognition systems. 
The major features of the palmprints are palm lines, wrinkles and ridges, and many approaches use them in different ways towards solving the palmprint recognition problem. Here we have proposed to use a set of statistical and wavelet-based features; statistical to capture the general characteristics of palmprints; and wavelet-based to find those information not evident in the spatial domain. Also we use two different classification approaches, minimum distance classifier scheme and weighted majority voting algorithm, to perform palmprint matching.
The proposed method is tested on a well-known palmprint dataset of 6000 samples and has shown an impressive accuracy rate of 99.65\%-100\% for most scenarios.
\end{abstract}


\maketitle

\section{Introduction}
{Identification} has always been required in critical tasks and applications; to ask for an object or a signature that only the right person possesses. Throughout history, there were always attempts to make this process flawless and secure, mostly to prevent forgeries. For centuries, identity was confirmed through an item or a mark.	Today there are many ways for a person to identify himself or herself, including passwords and keys. A very reliable way is to utilize something that is very difficult to duplicate quickly; features of the person himself, also known as biometric data. The latter began in the late 19th century with the collection of fingerprints for forensic purposes due to them being unique to every person from whom they are sampled. Afterwards many other characteristics were deemed efficient and unique to be used in the areas of security and identification. Various algorithms have been used on an individual's biometric data such as fingerprints \cite{finger}, iris patterns \cite{iris}, \cite{iris_mine}, face \cite{face} and palmprints \cite{palmprint}. Sometimes even several methods are used together and then cross-referenced to dramatically increase the verity of the judgment.

We chose palmprints to be our focus in this work, because we believe that despite their more simplicity than fingerprints which casts the illusion that their use is less secure, they can be utilized just as reliably. Palmprints are more economical in the sense of acquisition. They can be easily obtained using inexpensive CCD cameras. They also work in different conditions of weather and are typically time-independent. However, due to sampling limitations, lighting and other factors, they may pose problems like insufficient data due to unclear wrinkles or confusion due to poor image quality. This is the reason there are usually many different samples from every person in the database.


Like all biometric data, the key is to use image processing and, in many cases, machine learning approaches to extract distinct traits of every person, called features, by their samples and use the captured data for the next blocks of data to come. Being a popular area of research, there are many set of features and different approaches used for palmprint recognition \cite{palmprint}; however, two general approaches for palmprint recognition are the following:
\begin{enumerate}
\item	Transforming palmprints into another domain and extracting the features in the transform domain, which could be wavelet, Fourier, Gabor, etc. 

\item	Trying to extract principal lines and wrinkles and other geometrical characteristics as discriminants.

\end{enumerate}

There are many transform-based approaches. Li proposed Fourier-based features for palmprint recognition \cite{Li}. 
Wu \cite{wave} presented a wavelet-based approach for palmprint recognition. They used wavelet energy distribution as a discriminant for the recognition process. Ekinci \cite{PCA+GWR} proposed a Gabor wavelet representation approach followed by kernel PCA for palmprint recognition.
There are also several line-based approaches, since palm lines are among the most useful features of palmprints. Chen \cite{crease} proposed a recognition algorithm that primarily uses creases. They extract all creases from a palm and use them for palmprint matching. The main advantage of this algorithm is that it is rotation- and translation-invariant. Jia \cite{jia} used robust line orientation code for palmprint verification. 
A few groups used image coding methods for palmprint recognition, such as palm code, fusion code, competitive code, ordinal code \cite{code}.
A survey about palmprint recognition algorithms before 2009 is provided by Kong \cite{palmprint}. 

In the more recent works, in \cite{hol}, Jia proposed a new descriptor for palmprint recognition called histogram of oriented lines (HOL) which is inspired by the histogram of oriented gradients descriptors. The proposed descriptor has some robustness against small deformation and changes of illumination.
In \cite{texture}, Minaee proposed to use a set of textural features for palmprint recognition. In their work, a set of local texture features are derived for each palmprint and then weighted majority voting algorithm is used to perform recognition task. In \cite{mistani}, Mistani proposed an energy-based feature which results in a high accuracy for palmprint recognition.
In \cite{Xu}, Xu proposed a quaternion principal component analysis approach for multispectral palmprint recognition which achieved a high accuracy rate. 

In this work, we have used the palmprint database created by the Polytechnic University of Hong Kong (PolyU) \cite{database} which includes a set of 12 palmprint samples from 500 people under four distinct light spectra. The job of the identifier is to take the picture of a new palmprint sample called a test subject and determine the person in possession of the most similar palmprint.
Our dataset allows us to use multiple spectra of the same palmprint. Multispectral methods require different samples of the same object in order to make a better decision \cite{16}. The images in this dataset are preprocessed and the regions of interest (ROI) for each of them are extracted. As a result, no more preprocessing is required before feature extraction. Four different palmprint images are shown in Figure 1.
\begin{figure}[1 h]
\begin{center}
    \includegraphics [scale=0.6] {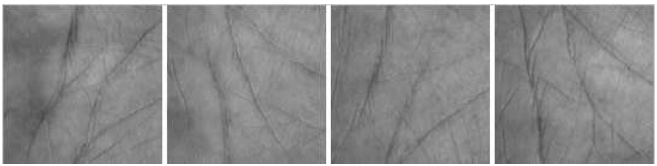}
\end{center}
  \caption{Four sample palmprints from PolyU database}
\end{figure}

Here we decided to use a set of features which capture the palmprint information both in spatial and frequency domains. We first divide each image into non-overlapping blocks and then extract 5 statistical features to capture essential spatial information and 9 wavelet-based features to determine the frequency content of the image. Since the statistical features alone are not able to capture high-frequency patterns in palm images, we also use wavelet features to capture fine details of palm images so we are able to detect the partial differences between two different palmprints.

After feature extraction, we have to use a classification algorithm to identify palmprints. In this work, two different classifiers are used, the first one being minimum distance classifier and the other one is the weighted majority voting algorithm, which is very fast and can be also implemented in electronic devices in conjunction with energy-efficient algorithms \cite{hosseini1}, \cite{hosseini2}.

The rest of the paper is organized as follows. Section \ref{SectionII} provides a detailed explanation of the proposed features. The minimum distance classifier and weighted majority voting algorithms are explained in Sections \ref{SubsectionIIIA} and \ref{SubsectionIIIB} respectively. Experimental results are given in Section \ref{SectionIV}. We have provided a comprehensive comparison with other state-of-the-art algorithms there. In the end, conclusion is given in Section \ref{SectionV}.

\section{Features}
\label{SectionII}

In general, features play a crucial part in the area of machine learning and computer vision. The more informative features are, the higher accuracy one can get. Therefore it is of utmost importance to extract a set of features which have the required information for prediction of the target value. Once the images are dealt with, it is usually needed to extract a set of features from them to use for prediction. For a comprehensive study of feature extraction, the reader is referred to \cite{feature}.

There are different kinds of features that can be used for palmprint recognition. One type consists of spatial and statistical features. Another type is transform-domain features such as Fourier, Wavelet and Gabor-based features. Another category is the geometrical features based on principal lines and wrinkles. This category requires to extract these lines from the palmprint first, which may not be very simple for low-resolution images. Foreground segmentation techniques can be used to extract principal lines from palmprint \cite{LAD}. Geometrical features are also used in other applications \cite{chromosome}. Sparsity-based features have also drawn a lot of attention in image classification during the past few years \cite{hojjat1}-\cite{hojjat3}.

Here a set of features is used to capture the behavior of the palmprint in both spatial and frequency domains. Based on the simulations, this results in a very highly accurate identification method for palmprints. Two images may have similar global characteristics but look different in local regions. Thus the local features are extracted from different parts of each palmprint and combined to create a feature matrix for every image.

Each palmprint is divided into non-overlapping blocks, and from each block, 5 statistical and 9 wavelet-based features are derived which are expected to determine the frequency information of the palms. To obtain the statistical features of each block, it is necessary to find the histogram of pixel intensities first. 

Let us assume that $B(i,j)$ represents the pixel value at the location $(i,j)$ of a block of size $N\times N$ (here $N=16$) and that $p(k)$ denotes the probability mass function for the $k$-th pixel value, $v(k)$, in that block. Now the 5 following attributes can be defined as the statistical features of the current block:
\begin{gather*}
\hspace{-1.1cm} f_1=E[v]= \sum_{k=1}^{K} p(k)v(k) \\
\hspace{-1.9cm}  f_2=E[(v-E[v])^2] \\
\hspace{-1.9cm}  f_3=E[(v-E[v])^3] \\
\hspace{-1.9cm}  f_4=E[(v-E[v])^4] \\
\hspace{0.97cm} f_5=Entropy(p)= - \sum_{k=1}^{K} p(k)\log_2 p(k)
\end{gather*}
where $K$ denotes the number of different pixel values in the current block. 

The other 9 features are wavelet-based. In this work, the wavelet transform used is the second-order Daubechies filter \cite{Daubechies}. The 2D-wavelet decomposition is performed up to three stages, and in the end, 10 subbands are produced. Since the mean pixel intensity is used as a statistical feature already, it is not required to use the LL subband of the last stage, but all other 9 subbands may be utilized. We extract the wavelet features in our implementation using the following algorithm:

\begin{enumerate}
\item	Divide each palm image into $N\times N $ non-overlapping blocks;

\item	Decompose each block up to 3 levels using Daubechies-2 wavelet transform; and

\item	Compute the energy of each subband and put the similar subband energy of all blocks in a vector.
\end{enumerate} 

If each subband is denoted by $d_i$ where $i=1,2,...,9$, the wavelet features can be derived as follows:
\begin{gather*}
f_{5+i}=E[d_i^2],  \ \ \ \ \ \ \ \  i=1,2,...,9
\end{gather*}

 Note that $d_1,d_2,d_3$ are blocks of size $\frac{N}{2} \times \frac{N}{2}$, $d_4,d_5,d_6$ are blocks of size $\frac{N}{4} \times \frac{N}{4}$ and $d_7,d_8,d_9$ are blocks of size $\frac{N}{8} \times \frac{N}{8}$. An example of 3-level wavelet decomposition of a palmprint is presented in Figure 2.

\begin{figure}[2 h]

\begin{center}
    \includegraphics [scale=0.7] {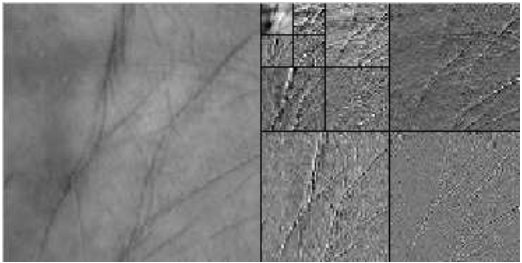}
\end{center}
  \caption{Left: A palm image, Right: 3-level wavelet decomposition of the image}
\end{figure}

After the computations, there will be 14 different features for each block which can be combined in a vector together: $\textbf{f}=(f_1,f_2,...,f_{14})^\intercal$. It is necessary to find the mentioned features for each block of a palmprint. If each palm image has a size of $S_W \times S_H$, the total number of non-overlapping blocks will be:
\begin{gather*}
M=\frac{S_W \times S_H}{N^2}
\end{gather*}
Therefore there are $M$ such feature vectors, $\textbf{f}^{(m)}$. Similarly they can be put in the columns of a 2-dimensional matrix to produce the feature matrix of that palmprint, $\textbf{F}$:
\begin{gather*}
\textbf{F}=[\textbf{f}^{(1)} ; \textbf{f}^{(2)}; ... ;\textbf{f}^{(M)}]
\end{gather*}
Therefore there will be a total number of $14\times\textbf{M}$ features for each palmprint image.

\section{Recognition algorithm}
\label{SectionIII}
The goal of palmprint recognition is to identify a person using their palmprint samples. It is possible to use the derived features of each person for identification. After finding the features of all people in the dataset, a classifier is required so that the features of each test palmprint can be compared with all of the available samples in the dataset and find the most similar one. There are different classifiers that can be used for this job; for example, minimum distance classifiers, support vector machines and probabilistic neural networks. In our work, two different classifiers are used. One is the minimum distance classifier which finds the most similar palmprint by minimizing a distance between the features of the test samples and those of the training samples. The other one is the weighted majority voting algorithm which finds the most similar palmprint by acquiring the predictions based on each feature and its weight, each time awarding the training data with points, and choosing the entry with the highest point. These two algorithms are described in the following sections. Since there are enough data in our dataset, our only goal is to minimize the recognition error on test samples, but if one is dealing with a small dataset, the over-fitting problem should also be considered, as it is discussed in \cite{mtbi}.

\subsection{Minimum Distance Classifier}
\label{SubsectionIIIA}
The minimum distance classifier is a popular algorithm in the template matching area. Basically, it finds the distance between the features of an unknown sample and those of the training samples and picks the training sample which has the minimum distance to the unknown as the predicted label.
Therefore if $F^{(t)}$ denotes the features of a test sample and $F^{(k)}$ denotes the features of the $k$-th sample in our dataset, minimum distance assigns the test sample to one of the samples in the dataset such that:
\begin{gather*}
k^*=\argmin_k \big[ dis(F^{(t)},  F^{(k)}) \big]
\end{gather*}
Here Euclidean distance is used, which results in the nearest neighbor classifier. 

In this algorithm, the feature matrix of all palmprints are extracted first. Considering size of the image and the block, each feature matrix has a size of $14\times64$.

As previously mentioned, there are 500 different persons in the database, and for each, there are 12 sample images. Every time, $M$ of these 12 samples are assigned as training and the remaining ones ($12-M$) as test samples, leading to a total of $500(12-M)$ test samples. For each person, the feature matrix is defined as the average of the feature matrices of the $M$ different training images of that person. Then, for an unknown sample with the feature matrix $F^{(t)}$, the following distance should be found:
\begin{gather*}
dis(F^{(t)},F^{k})= \sum_{i=1}^{14} \sum_{j=1}^{64} w_i \alpha_i (F^{(t)}_{ij}-F^{(k)}_{ij})^2
\end{gather*}
which is very similar to the Frobenius norm of the difference of the two matrices, and each row has a weight of $w_i\alpha_i$, where $\alpha_i$ is a feature-normalizing factor trying to map all features into the same range. The term $\alpha_i$ can be defined as the reciprocal of the mean value of the corresponding feature of all training samples. $w_i$ is the feature importance factor which gives higher weight to the features with more information about image labels. This factor can be any increasing function of single feature accuracy. Here $w_i$ is defined as the recognition accuracy when the $i$-th row of the feature matrix is used on its own for the recognition process.

For each palmprint, there are four different spectra; red, green, blue and infrared. Their features are signified by $F_j^{(r)}$, $F_j^{(g)}$, $F_j^{(b)}$ and $F_j^{(i)}$ respectively. The key is to calculate the above distance for all the spectra by comparing the images in the same spectrum. Next, the distance between a test image and the $k$-th training sample will be defined as the average of the distances of their corresponding spectra. Then, the predicted entry for a test image with the feature matrix $F^{(t)}$ will be:
\begin{gather*}
k^*=\argmin_k \big[ dis(F^{(t)},  F^{(k)}) \big]
\end{gather*}

\subsection{Weighted Majority Voting}
\label{SubsectionIIIB}

Voting theory has many applications in AI, search engines and recommendation systems. In algorithms based on majority voting, every voter decides the outcome of the test on its own, and in the end, all the decisions are counted and the final verdict is given. Here the voters are the features and the votes are given to every person in the training samples. In the unweighted case, all features have the same impact on the votes and none of them is superior. In the weighted case, which is used here, each feature has a weight of its own, based on which points will be awarded to each person. When added, the score will decide to which profile the test image is the most analogous.

This scheme has a very simple algorithm and can be performed in a very short time compared to other works in this field. 
First, the images of every single person are uniformly shuffled in the database so that the  training part can use different pieces of data from a random set of the 12 images. Then, the features of the all the training data are gathered and the feature average for every person is computed. Next, the other images are used as test subjects and, for every existing spectrum, the distance between the feature vector of every sample and the average matrix from the training period is calculated. 
The minimum distance with any subject based on every feature is awarded points based on the coefficient of the feature in that stage. This reward is also applied to a matrix shared by all four spectra and holds the total score. In the end, the person gaining the maximum of the global score matrix is identified as the answer to the recognition query.

For every feature vector $\textbf{f}_i$, the voting result will be: 
$$k^*(i)=\argmin_k||\textbf{f}_i^{(t)}-\textbf{f}_i^{(k)}||_2$$
When $\textbf{f}_i$ finds the person with minimum distance to the test subject, that person receives a point equal to the weight of the feature. 

The score of person $j$ based on $\textbf{f}_i$ is denoted by $w_iS_j(i)$ where $w_i$ is the weight of the corresponding feature and $S_j(i)=I(j=\argmin_k|\textbf{f}_i^{(t)}-\textbf{f}_i^{(k)}|)$ where $I(x)$ is the indicator function. Then the total score of the $j$-th training sample based on all the features in the scope of all the colors will be:
$$S_j=\sum_{All~colors}\sum_{i=1}^{i_{max}}{w_iI(j=\argmin_k|\textbf{f}_i^{(t)}-\textbf{f}_i^{(k)}|)}$$

In the end, the identification factor $j^*$ will be calculated:

\begin{gather*} 
j^*=\argmax_j \big[ S_j\big]= \argmax_j \big[\sum_{All~colors}\sum_{i}w_iS_j(i) \big]
\end{gather*}

\section{Results}
\label{SectionIV}

We have tested our algorithm on the PolyU multisprectral palmprint database \cite{database} containing 6000 palmprints captured from 500 different palms. Every palm is sampled 12 times in two sessions. Each palmprint contains 4 palm images collected at the same scene under 4 different illuminations, including red, green, blue and NIR (near-infrared). Therefore the total number of images is 24000. The resolution of each image is 128$\times$128.
As mentioned before, we are working on preprocessed palmprint images. Therefore, no further action is required to align or resize the palm images.

We have performed palmprint recognition for different fractions of training and test images. Correct identification takes place when the test palmprint is classified as a person whose label is the same as the label of this palmprint, and misidentification occurs when the test palmprint is classified for an entry whose tag is different from that of the correct palmprint.

Table \ref{TblRes1} denotes the identification accuracy for two different classifiers. Every result is produced by repeating the experiment 10 times and taking the average of their results in order to make it more precise.

\begin{table} [h]
  \caption{Identification accuracy for minimum distance classifier and weighted majority voting algorithm}
  \centering
\begin{tabular}{|m{2.3cm}|m{2.5cm}|m{2.4cm}|}
\hline
Ratio of training samples & Using minimum distance classifier & Using weighted majority voting  \\
\hline
\ \ \ \ \ \ 3/12 & \ \ \ \ \ \ 97.42 & \ \ \ \ \ \ \ 99.95  \\
\hline
\ \ \ \ \ \ 4/12 & \ \ \ \ \ \ 99.72 & \ \ \ \ \ \ \ 99.99  \\
\hline
\ \ \ \ \ \ 5/12 & \ \ \ \ \ \ 99.51 & \ \ \ \ \ \ \ 99.99  \\
\hline
\ \ \ \ \ \ 6/12 & \ \ \ \ \ \ 100 & \ \ \ \ \ \ \ 99.99  \\
\hline
\end{tabular}
\label{TblRes1}
\end{table}

It can be seen that when using a lower number of training samples, weighted majority voting fares much better that minimum-distance classifier. The reason is the fact that there are many features deciding the output of our system, and if a group of them fail to successfully pinpoint the match to the test subject, there are still others to help the system find the correct entry.

Table \ref{TblComp} shows a comparison of the results of our work and those of five other highly accurate schemes. We have compared our work with methods which were introduced in recent years. K-PCA+GWR denotes Ekinci's approach which applies kernel PCA to the Gabor features \cite{PCA+GWR}. MDA+GWR denotes multilinear discriminant analysis applied to Gabor representation which is presented in \cite{hol}. The reported accuracy of the proposed scheme in Table 2 corresponds to the case where half of the images (3000 multispectral images) of each person are used for training and the other half for testing.
For more details about the experiment conditions of other works, the reader is referred to the referenced papers in the first column of Table 2.

\begin{table} [h]
\centering
  \caption{Comparison with other algorithms for palmprint recognition }
  \centering
\begin{tabular}{|m{6cm}|m{2cm}|}
\hline
\ \ \ \ \ \ \ \ \ \ \ \ \ \ \ \ \ \ \ \ \ \ \ \ \ Method &  Recognition rate\\
\hline
K-PCA+GWR \cite{PCA+GWR} & \ \ \ \ \ 95.17\% \\
\hline
Quaternion principal component analysis \cite{Xu} & \ \ \ \ \ 98.13\% \\
\hline
MDA+GWR \cite{hol} & \ \ \ \ \ 98.81\% \\
\hline
Histogram of Oriented Lines \cite{hol} & \ \ \ \ \  99.97\% \\
\hline
Textural features \cite{texture} & \ \ \ \ \  100\% \\
\hline
Proposed scheme using majority voting & \ \ \ \ \  99.99\% \\
\hline
Proposed scheme using minimum distance classifier & \ \ \ \ \  100\% \\
\hline
\end{tabular}
\label{TblComp}
\end{table}

As it can be seen, the algorithm utilized in this paper outperforms the other methods. This can be due to the fact that statistical features are also used in parallel with wavelet-based ones. It is known that wavelet transform is quite sensitive to small changes in the image due to deformation, distortion and other transformations. As a result, methods solely based on such features are more susceptible to noise and other distortions. However, the proposed statistical features in this work do not share this drawback. Therefore they can help to have a more accurate recognition algorithm. 

The system is implemented using MATLAB on a laptop with Windows 7 and Core i7 CPU running at 2GHz. The execution time for the proposed method is about 0.05s per test using majority voting algorithm.

\section{Conclusion}
\label{SectionV}

This paper proposed a set of statistical and wavelet-based features for palmprint recognition. One attempts to find the spatial information of palm images and the other aims to mostly capture their frequency content. One is sensitive to the major difference between different palms, while the other is more perceptive of the partial differences between similar palmprints. Two different classifiers are used to perform the recognition process. By using this method, our algorithm is able to identify palmprints with similar line patterns as well as unclear palmprints.

The proposed algorithm has significant advantages over the previous popular methods. The used features are very simple to extract. The algorithm is very fast and it does not need classifier training. Most importantly, it has a very high accuracy rate which is robust to the number of training samples and can be high even for the case where the ratio of training to test is 1 to 3.
In the future, we will apply this set of features to other biometrics as well.

\section*{Acknowledgments}
The authors would like to thank the Hong Kong Polytechnic University (PolyU) for sharing their multisprectral palmprint database with us \cite{database}.

\end{document}